# Precisely Predicting Acute Kidney Injury with Convolutional Neural Network Based on Electronic Health Record Data

Yu Wang [a]    JunPeng Bao [a*]    JianQiang Du [b]    YongFeng Li [c]

[a] The School of Computer Science and Technology, Xi'an Jiaotong University, Xi'an, P.R.China

[b] The School of Life Science and Technology, Xi'an Jiaotong University, Xi'an, P.R.China

[c] Information Department, Lishui People's Hospital, Lishui, P.R.China

[*] Corresponding author, Doctor, Tel.: +86-13186003037, Fax numbers: +86-029-82668971, baojp@mail.xjtu.edu.cn,

Mailing Address: No.28, Western Xianning Road, BeiLin District, Xi'an 710049, P.R.China

## Abstract

**Background:** The incidence of Acute Kidney Injury (AKI) commonly happens in the Intensive Care Unit (ICU) patients, especially in the adults, which is an independent risk factor affecting short-term and long-term mortality. Though researchers in recent years highlight the early prediction of AKI, the performance of existing models are not precise enough.

**Objectives:** The objective of this research is to precisely predict AKI by means of Convolutional Neural Network on Electronic Health Record (EHR) data. The data sets used in this research are two public Electronic Health Record (EHR) databases: MIMIC-III and eICU database.

**Methods:** In this study, we take several Convolutional Neural Network models to train and test our AKI predictor, which can precisely predict whether a certain patient will suffer from AKI after admission in ICU according to the last measurements of the 16 blood gas and demographic features. The research is based on Kidney Disease Improving Global Outcomes (KDIGO) criteria for AKI definition.

**Results:** Our work greatly improves the AKI prediction precision, and the best AUROC is up to 0.988 on MIMIC-III data set and 0.936 on eICU data set, both of which outperform the state-of-art predictors. And the dimension of the input vector used in this predictor is much fewer than that used in other existing researches.

**Conclusion:** Compared with the existing AKI predictors, the predictor in this work greatly improves the precision of early prediction of AKI by using the Convolutional Neural Network architecture and a more concise input vector. Early and precise prediction of AKI will bring much benefit to the decision of treatment, so it is believed that our work is a very helpful clinical application.

# 1. Introduction

The incidence of Acute Kidney Injury (AKI) commonly happens in ICU patients, especially in adults, which is an independent risk factor affecting short-term and long-term mortality[1, 2, 3, 4]. So, early prediction of AKI is highlighted by researchers in recent years.

Electronic Health Records (EHR) provides an opportunity to predict the severity of diseases, such as AKI. Lots of the existing studies on the early prediction of AKI use traditional machine learning methods. Li Y et al.[5] demonstrate a logistic regression method based on word embedding features of clinic notes to predict AKI onset after ICU admission. Mohamadlou H et al.[6] use several lab measurements and decision tree method to predict AKI onset after admission in ICU. Zimmerman L et al.[7] adopts logistic regression and random forest (RF) methods to predict AKI after ICU admission. The best AUROC of these methods is only 0.796. A few existing reports about the early prediction of AKI use deep learning methods. Pan Z et al.[8] propose a recurrent neural network (RNN) based method to predict the onset of AKI in ICU. Li Y et al.[5] also proposed a word embedding based convolutional neural network (WB-CNN). However, the best AUROC of these two work are just 0.893 and 0.738 respectively. It is obvious that the performance of the state-of-art is not precise enough. The best AUROC of these researches is only 0.893 on MIMIC-III data set and 0.871 on eICU data set. Moreover, most of the aforementioned researches take as many variables as possible into consideration.

This paper takes full advantages of convolutional neural network (CNN) models to precisely predict the onset of AKI. The inputs of the predictor are 16 features of blood gas measurements and demographics, which are much fewer than the features used in the existing predictors. This study is based on two public available EHR data sets and Kidney Disease Improving Global Outcomes (KDIGO) criterion[9] to definite AKI. Finally, our predictor greatly improves the prediction precision of AKI, and the best AUROC is up to 0.988 on MIMIC-III data set and 0.936 on eICU data set, which outperforms all of the state-of-the-art predictors[5, 6, 7, 8]. The source code can be accessed online[1].

## 2. Objectives

The objective of this research is to precisely predict whether a certain patient will suffer from AKI after admission in ICU according to the last measurements of the features before admission. The data sets used in this paper are drawn from Medical Information Mart for Intensive Care III (MIMIC-III)[10] and Phillips eICU Collaborative Research database[11]. The MIMIC-III database, consisting of 61,532 critical care patients in ICU, is a freely accessible critical care database with demographics, vital signs, laboratory tests, medications, etc. and is compiled by the MIT Lab for Computational Physiology. The eICU database, containing over 200,000 patient stays, is an openly available multi-center critical care database developed by Philips Healthcare in partnership with the MIT Lab for Computational Physiology.

## 3. Methods

### 3.1 Data processing

3.1.1 Input Vector

According to the relative studies[12, 13], AKI is relevant to some biochemistry measurements, such as blood gas measurements. We select blood gas features and fundamental demographics to make our input vector. But some blood gas features which

---
[1] https://github.com/Sophiaaaaaa/AKI-Prediction.

present in fewer than 20% patients are removed from the data set to avoid biases and noise. The final input vector can be represented as follows:

*[gender, age, weight, height, BMI group, bicarbonate, chloride, creatinine, glucose, magnesium, postassium, sodium, urea nitrogen, hemoglobin, platelet count, white blood cells]*.

Notably, urea nitrogen is excluded on eICU data set as most patients have no measurement of it.

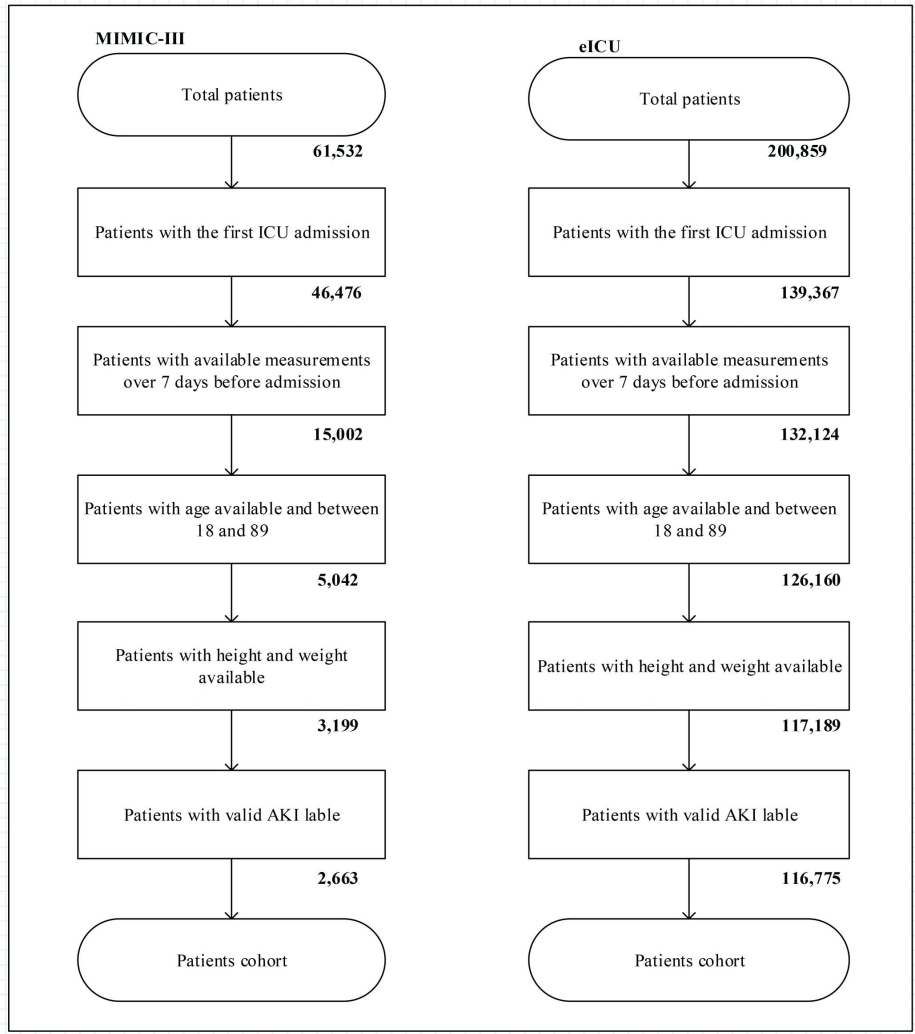

**Fig. 1.** The procedures of cohort creation on MIMIC-III data set and eICU data set.

3.1.2 Inclusion Criteria

Our target is to predict whether certain person will suffer from AKI after their admission in ICU. The features in the aforementioned input vector are represented by the last measurements during the seven days before the patients entered into ICU. Our research only considers the first ICU stays records of each patient, whose age are between 18 and 89 and whose height and weight are available. Labels for all the patients are based on the SQL scripts which were made by MIT Lab and openly available online[14]. Our procedures of cohort creation on the two databases are shown as *Fig. 1*. A total of 2,663 patients meet the inclusion criterion in MIMIC-III database and 116,775 patients are included in eICU database.

3.1.3 Imbalanced Data

After executing the inclusion criterion above, the MIMIC-III data set contains 2,505 positive and 158 negative examples, and the eICU data set contains 93,149 positive and 23,626 negative examples. However, classifying directly on these imbalance data may produces an inveracious predictor because the classifier often learns to simply predict the majority class. A lot of studies have proposed different algorithms to process imbalanced data in classification tasks. Haixiang G et al.[15] and Buda M et al.[16] have reviewed the existing approaches of imbalanced data classification, which can be divided into data-level methods and algorithm-level methods. Two strategies on data-level are resampling and feature selection. Algorithm-level methods include cost-sensitive learning[17] and ensemble methods[18]. Resampling strategies are more versatile because of their independence of the selected classifier[15]. It contains oversampling and undersampling techniques. Buda M et al.[16] has proved that oversampling gains better performance than undersampling and other algorithm-level methods. Synthetic minority oversampling technique (SMOTE)[19] is a more advanced oversampling method, which has been successfully used in many other research[20]. So this study uses the *imlearn* package for SMOTE, which interpolates among existing minority class examples and generates new minority class samples.

Table 1. The contents of the preprocessed data in this study.

| Data set Name | The number of the samples | The input vector of each sample |
| --- | --- | --- |
| MIMIC-III | 5010 (positive and negative samples are half and half) | [gender, age, weight, height, BMI group, bicarbonate, chloride, creatinine, glucose, magnesium, postassium, sodium, urea nitrogen, hemoglobin, platelet count, white blood cells] |
| eICU | 186,298 (positive and negative samples are half and half) | [gender, age, weight, height, BMI group, bicarbonate, chloride, creatinine, glucose, magnesium, postassium, sodium, hemoglobin, platelet count, white blood cells] |

After upsampling, a total of 5,010 samples (positive and negative samples are half and half) are formed for MIMIC-III database and 186,298 samples (positive and negative samples are half and half) for eICU database. *Table 1* describes the detailed information of the preprocessed data sets and these data sets are also available online[2]. We train and test on the two data sets respectively.

## 3 .2 Deep Convolution Neural Network

Convolutional Neural Network (CNN)[21, 22] is one of the supervised deep learning architectures, which contains several convolutional layers followed by one or more fully connection layers. CNN has shown its merits in many fields, such as computer vision, bioinformatics[23, 24], medical image[25, 26]. It extracts local features from the input data using convolutional layers, which uses the extracted features of the previous layer as input and provides a higher-level feature abstraction.

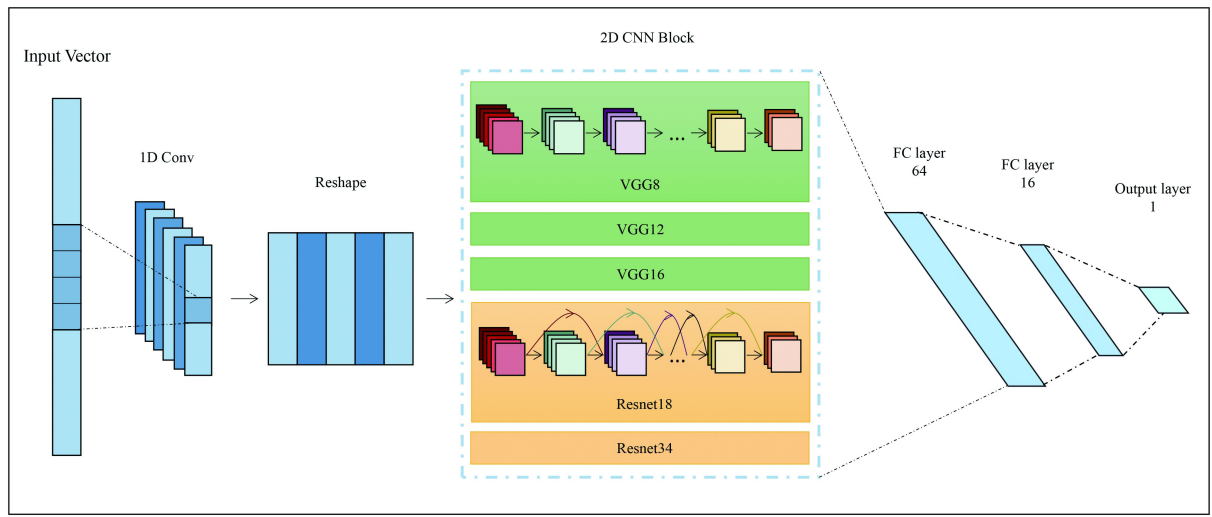

**Fig. 2.** The construction of CNN used in this study.

This study implements a predictor based on CNN to precisely predict AKI status. The input data is a 16-dim vector for MIMIC and 15-dim for eICU (details shown as *section 3.1.1* and *Table 1*). *Fig. 2* illustrates the architecture of our model, which contains an 1-dim convolutional layer, a reshape layer, 2-dim CNN block and two fully connection (FC) layers. The first layer is an 1-dim convolutional layer which consists of 16 filters. The filter size and the stride are 3 and 1 respectively. The following layer is a reshape layer which aims to transform 1-dim input to 2-dim images. Next is a 2-d CNN block in which we examine several CNN architectures, including VGG[27] with 8 layers, 12 layers, 16 layers and Resnet[28] with 18 layers, 34 layers. For each layer in 2-d CNN block, the filter size is set to $3 \times 3$ and the stride is set to 1. Finally, two full connection layers which contain 64 and 16 filters respectively and sigmoid function are designed to map the outputs of the model into prediction results.

---
[2] https://github.com/Sophiaaaaaa/AKI-Prediction/tree/master/data

Each architecture that we examine contains a batch normalization layer before 2-d convolution operations and a dropout layer between the two FC layers to alleviate over-fitting and bias. Dropout is applied to randomly select neurons to be ignored during the training stage, which means at each iteration, several neurons are set to 0, the other neurons still work together to train the neural network. Notably, dropout rate is set to 0.5 in our study. Batch normalization is a technique to limit covariate shift by normalizing the activation of each layer, which supposedly allows each layer to learn on a more stable distribution of inputs.

We use 5-fold cross-validation to train our models on the MIMIC-III and eICU data sets. Cross-validation is an efficient approach to ensure the generalization ability of the model. And 5-fold cross-validation means that the data set is randomly divided into fifth then forth of the fifths are used for training while the rest for testing, and this process is repeated so that all possible combinations of fifths are contained in the training period. We used the area under the receiver operating characteristic (AUROC) to indicate the performance of our models.

### 3.3 Ethical Considerations

This research is performed in compliance with the World Medical Association Declaration of Helsinki on Ethical Principles for Medical Research Involving Human Subjects. And both of the two databases used in this study have hidden the identification of each patient.

## 4. Result

Our research adopts Convolutional Neural Network to predict AKI status of each patient in ICU, and the predictor achieves the best AUROC up to 0.988 on MIMIC-III data set. This predictor is also trained and tested on eICU data set, and the best AUROC is up to 0.936.

We first compare our method with another deep neural network architecture named MLP by using the same input features. *Table 2* shows the results on both of the two data sets. Because of the gradient vanishing, MLP gets a degradation of the performance as the network goes deeper. On the contrast, our method shows much better performance than MLP whatever the number of layers is.

**Table 2.** The best AUROC of our method and MLP baseline on the two data sets.

| The Number of Layers | MIMIC-III data set | | eICU data set | |
| :---: | :---: | :---: | :---: | :---: |
|  | MLP | Our Method | MLP | Our Method |
| 8 | 0.852 | **0.974** | 0.676 | **0.834** |
| 12 | 0.844 | **0.986** | 0.674 | **0.899** |
| 16 | 0.840 | **0.975** | 0.692 | **0.916** |
| 18 | 0.837 | **0.988** | 0.673 | **0.925** |
| 34 | 0.667 | **0.984** | N/A* | **0.936** |

* The number of layers is too deep to get a valid result for MLP.

This work also compares the performance of our predictor with that of the existing predictors mentioned in *Section 1*, including those using logistic regression method[5, 7], decision tree (DT) based method[6], random forest (RF)[7], WB-CNN[5] and RNN[8]. *Table 3* shows the comparison of the performance of our predictor with that of the aforementioned predictors on MIMIC-III dataset. Beyond that, compared with the performance of the other predictor[8] on eICU data set (the best AUROC is 0.893 only), our predictor shows markedly improvement (the best AUROC is up to 0.936).

**Table 3.** Comparison of our AKI predictor with existing predictors on MIMIC-III dataset.

| Predictor | Description | The best reported AUROC |
|---|---|---|
| LR[7] | Using 32 lab results as input vector and Logistic Regression(LR) method, which is suitable to linear mapping. | 0.783 |
| L2-LR[5] | Using embedded clinical notes as input vector and Logistic Regression with L2 normalization (L2-LR) method. L2 normalization can improve generalization ability. | 0.779 |
| DT[6] | Using 6 lab results as input vector and XGBoost method, which ensembles multiple Decision Trees (DT) to improve the performance. | 0.878 |
| RF[7] | Using 32 lab results as input vector and Random Forest (RF) model. Random Forest is another ensemble decision tree model, which corrects for decision trees' habit of overfitting. | 0.779 |
| WB-CNN[5] | Using embedded clinical notes as input vector and CNN model, which is more suitable for large scale data sets with short texts. | 0.738 |
| RNN[8] | Using a 51-dimension input vector and a self-correcting RNN. RNN is suitable for time series data but it costs much time. | 0.893 |
| Our Method | Using 16 kinds of lab measurements as input vector and several deep convolutional neural network models. This work outperforms all of the existing AKI predictors with the performance much closed to the best. | **0.988** |

## 5. Discussion

This work provides a method to precisely predict AKI among adult patients in ICU and greatly improves the AKI prediction precision. The best AUROC up to 0.988 on MIMIC-III data set and 0.936 on eICU data set, which are both the highest results in the field of AKI prediction. In particular, this study uses much fewer blood gas measurements and demographics as the input vector than the existing predictors, most of which use as many available measurements as they can obtain, including vital signs, lab values, demographic features, etc.. *Table 4* compares the input vector of this predictor with the existing predictors.

**Table 4.** Comparison of the input vector of the existing predictors with that of ours.

| Authors | The Length of Input Vector | Features Included in the Input Vector |
|---|---|---|
| Li Y et al.[5] | Unknown* | patients age, gender, race/ethnicity, clinical notes during the first 24 hours of ICU admission (preprocessed by word embedding), and 72-hour serum creatinine after admission. |
| Mohamadlou H et al.[6] | 6 | heart rate, respiratory rate, temperature, SCr, Glasgow Coma Scale (GCS), and patient's age |
| Zimmerman L et al.[7] | 32 | patient age, gender, ethnicity, 72-h creatinine, and 28 vital signs and lab values in total. |
| Pan Z et al.[8] | 51 | age, gender, 24 co-morbidities, 4 vital signs, 16 lab measurements, Urine Output, Fluid Balance and 4 interventions |
| Our Method | 16 on MIMIC-III and 15 on eICU | gender, age, weight, height, BMI group, bicarbonate, chloride, creatinine, glucose, magnesium, postassium, sodium, urea nitrogen, hemoglobin, platelet count, white blood cells |

* The author did not declare the length of input vector they used in their predictor.

Imbalanced data commonly occurs in real world, especially in the area of clinical. This paper adopts one of the most widely used algorithm SMOTE to balance these data. However, there are much more methods being proposed in the recent years (described in *Section 3.1.3*). So in the subsequent studies, much work should be done to explore a more adaptable balanced method in dealing with clinical data.

In the future, we will pay more attention on feature selection, by which we can analyze which feature is the most significant to predict AKI and whether there are some features that can be included or excluded. Furthermore, this study focuses on binary classification, which tells whether certain patient will suffer from AKI or not after admission in ICU. More detailed classification can also be done under some specialized knowledge. Although some existing research predicted AKI in certain time period, such as in 24 hour after admission, the performance of their models are not precise enough. So in the future work, it is significant to explore AKI prediction in certain time period for clinical utility. More trials are needed to further evaluate the clinical utility of this approach to identify the at-risk patients.

## 6. Conclusions

This paper provides a very effective method to precisely predict AKI after admission in ICU. Moreover, the input vector of the predictor are more concise. The trained predictor can work efficiently, and it can be used to predict new AKI onset with a competitive AUROC (the best AUROC 0.988 on MIMIC-III data set and 0.936 on eICU data set). However, there are still a lot of work can be done in the future. We suggest that the prospective trials are needed to further evaluate the clinical utility of our approach on other data sets, and further studies should focus on feature selection and more detailed prediction.

**Conflict of interest**

The authors declare that they have no conflict of interest in this research.